\documentclass[pdflatex,sn-mathphys-num]{sn-jnl}% Math and Physical Sciences Numbered Reference Style 
\usepackage{graphicx}%
\usepackage{multirow}%
\usepackage{amsmath,amssymb,amsfonts}%
\usepackage{amsthm}%
\usepackage{mathrsfs}%
\usepackage[title]{appendix}%
\usepackage{xcolor}%
\usepackage{textcomp}%
\usepackage{manyfoot}%
\usepackage{booktabs}%
\usepackage{algorithm}%
\usepackage{algorithmicx}%
\usepackage{algpseudocode}%
\usepackage{listings}%

\theoremstyle{thmstyleone}%
\theoremstyle{thmstyletwo}%

\theoremstyle{thmstylethree}%

\begin{document}

\title[Article Title]{Analyzing Nobel Prize Literature with Large Language Models}
\author{Zhenyuan Yang\textsuperscript{*}$^{1}$, Zhengliang Liu\textsuperscript{*}$^{2}$, Jing Zhang\textsuperscript{*}$^{3}$, Cen Lu$^{4}$, Jiaxin Tai$^{5}$, Tianyang Zhong$^{6}$, Yiwei Li$^{2}$, Siyan Zhao$^{7}$, Teng Yao$^{8}$, Qing Liu$^{9}$, Jinlin Yang$^{8}$, Qixin Liu$^{9}$, Zhaowei Li$^{9}$, Kexin Wang$^{9}$, Longjun Ma$^{9}$, Dajiang Zhu$^{3}$, Yudan Ren$^{9}$, Bao Ge$^{8}$, Wei Zhang$^{10}$, Ning Qiang$^{8}$, Tuo Zhang$^{6}$, Tianming Liu\textsuperscript{\textdagger}$^{2}$
 % \#Corresponding Author % <-this % stops a space
 \affil[1]{Guanghua School of Management, Peking University, Beijing, China}
\affil[2]{University of Georgia, Athens, GA, USA}
\affil[3]{Computer Science and Engineering, The University of Texas at Arlington, Arlington, TX, USA}
\affil[4]{College of Computing, Georgia Institute of Technology, Atlanta, USA}
\affil[5]{Department of communication engineering, Shaanxi Normal University, Xi'an, China}
\affil[6]{School of Automation, Northwestern Polytechnical University, Xi’an, China}
\affil[7]{Xi’an University , Xi'an, China}
\affil[8]{School of Physics and Information Technology, Shaanxi Normal University, Xi'an, China}
\affil[9]{School of Information Science and Technology, Northwest University Xi'an, China}
\affil[10]{School of Computer and Cyber Sciences, Augusta University, Augusta, GA, USA}
}

% \author[1,2]{\textbf{A}$^{\ast}$\thanks{Equal contribution.}}
% \author[2]{\textbf{B}$^{\ast}$}
% \author[1,2]{\textbf{C}}
% \author[1]{\textbf{D}}
% \author[1]{\textbf{E}}
% \author[2]{\textbf{F}}
% \author[2]{\textbf{G}}
% \author[1]{\textbf{H}}
% \author[1]{\textbf{I}\thanks{Corresponding author: }}
% \author[2]{\textbf{J}$^{\dagger}$}
% \author[2]{\textbf{K}$^{\dagger}$}

% \affil[1]{University of Georgia, Athens, GA, USA}
% \affil[2]{Your Other Institution, Location}

\date{}
% Co-first authors
% \author{\textbf{Y}$^{1,2}$\thanks{Equal contribution} , \textbf{L}$^2$\footnotemark[1] ,\textbf{ Zihao Wu}$^{1,2}$, \textbf{Hanqi Jiang}$^1$, \textbf{Yi Pan}$^1$
% \textbf{Pengfei Jin}$^2$, \textbf{Sifan Song}$^2$, \textbf{Yucheng Shi}$^1$,  
% \textbf{Tianming Liu}$^1$\thanks{Corresponding author email: xli60@mgh.harvard.edu} , \textbf{Quanzheng Li}$^2$\footnotemark[2] , \textbf{Xiang Li}$^2$\footnotemark[2] \\
% ${}^1$ University of Georgia, 
% ${}^2$\\
% ${}^*$Equal Contribution.\\
% ${}^{\dagger}$ Correspodance Author.

% }



% \affil[1]{Department of Mathematical and Statistical Sciences, University of Alberta, Edmonton, Canada}


\abstract{This study examines the capabilities of advanced Large Language Models (LLMs), particularly the o1 model, in the context of literary analysis. The outputs of these models are compared directly to those produced by graduate-level human participants. By focusing on two Nobel Prize-winning short stories, 'Nine Chapters' by Han Kang, the 2024 laureate, and 'Friendship' by Jon Fosse, the 2023 laureate, the research explores the extent to which AI can engage with complex literary elements such as thematic analysis, intertextuality, cultural and historical contexts, linguistic and structural innovations, and character development. Given the Nobel Prize's prestige and its emphasis on cultural, historical, and linguistic richness, applying LLMs to these works provides a deeper understanding of both human and AI approaches to interpretation. The study uses qualitative and quantitative evaluations of coherence, creativity, and fidelity to the text, revealing the strengths and limitations of AI in tasks typically reserved for human expertise. While LLMs demonstrate strong analytical capabilities, particularly in structured tasks, they often fall short in emotional nuance and coherence, areas where human interpretation excels. This research underscores the potential for human-AI collaboration in the humanities, opening new opportunities in literary studies and beyond.}



\maketitle
\footnotetext[1]{\textsuperscript{*}Equal Contribution.}
\footnotetext[2]{\emph{Corresponding author: Tianming Liu (email: tliu@uga.edu).}}

% \thanks{\textsuperscript{*}Equal Contribution.}%
% \thanks{\emph{Corresponding author: Tianming Liu (email: tliu@uga.edu).}}
\section{Introduction}\label{sec1}

Since the inception of artificial intelligence (AI) in the 1950s, researchers have been driven by one central question: how well can AI perform in creative tasks? Whether it is generating art, composing music, or analyzing literature, the goal of understanding and replicating human creativity through machines has long been a key focus \cite{cardoso2009converging, shanahan2023evaluating}. Philosopher Ludwig Wittgenstein once said, “The limits of my language mean the limits of my world.” This idea resonates powerfully in today’s era of advanced AI, where breakthroughs in Large Language Models (LLMs), such as GPT-4 \cite{openai2024gpt}, Gemini \cite{team2023gemini}, Llama \cite{touvron2023llama}, and OpenAI’s newly released o1 model seem to bring us closer to Artificial General Intelligence (AGI) \cite{zhong2024evaluation}. These models trained on neural networks with billions of parameters and vast corpora of natural language, offer unprecedented capabilities in text generation, comprehension, and analysis. Their impressive abilities have drawn significant attention to their potential in natural language understanding and generation.  But to what extent can AI touch the boundaries of human thought? Exploring this question is also an exploration of how far we can replicate human intelligence.

As LLMs extend their applications far beyond traditional NLP tasks, they open new pathways in creative and intellectual domains that were once thought to be uniquely human. In particular, their potential to contribute to fields like literary analysis has sparked growing interest \cite{zhao2024understanding, nikolova2024contemporary, michel2024realistic,shanahan2023evaluating, yu2024lfed}. While LLMs have demonstrated utility in structured tasks like machine translation \cite{zhang2024binarized} and text generation \cite{li2024pre}, their effectiveness in more nuanced areas—such as interpreting and critiquing literature—remains less explored.

This research aims to bridge that gap by evaluating how well an advanced LLM can perform in literary analysis compared to human participants, with a focus on works by Nobel Prize-winning authors. Specifically, our study centers around two short stories, "Nine Chapters" by Han Kang and "Friendship" by Jon Fosse. The study examines the o1 model’s ability to engage with complex literary dimensions, including: thematic analysis; intertextuality and literary influence; cultural and historical contexts; linguistic and structural innovations; character development; moral and philosophical insights; narrative techniques and temporal structure; and emotional tone and psychological depth. By comparing the model's outputs with those of graduate-level human participants, this investigation seeks to assess the strengths and limitations of AI in literary interpretation, exploring whether AI can rival or complement human analysis in this traditionally humanistic field.

As AI continues to permeate creative and intellectual domains, understanding its capabilities in tasks like literary analysis will be crucial for shaping future collaborations between human and machine intelligence. Through a combination of qualitative and quantitative evaluations, this study contributes to the broader discourse on the evolving role of AI in the humanities. Evaluating the creativity of the advanced AI models is crucial for exploring their potential and limitations, ultimately pushing technological advancements forward.

\section{Related Work}

As LLMs continue to evolve, their integration into different domains has sparked significant academic interest. To understand the impact of LLMs requires a comprehensive examination of recent advancements and applications. We explore some existing work that underscores the multifaceted roles of LLMs in text analysis and literary evaluation.

\paragraph{Applications of LLMs in Text Analysis} 

Recent research emphasizes the expanding role of LLMs in text analysis and evaluates their capabilities inn various domains such as thematic analysis, translation, creative writing, and qualitative research \cite{Kaddour2023Challenges,Wang2023Document-Level,Chakrabarty2023Art}. Notably, studies have utilized LLMs for thematic analysis \cite{DBLP:conf/emnlp/DaiXK}, and extending this research to languages beyond English \cite{DBLP:journals/corr/abs-2404-08488}. Additionally, LLMs have also been applied for thematic analysis in more specialized fields like legal studies \cite{DBLP:conf/jurix/DrapalWS23} and poetry \cite{Poetry}, as well as in specific formats such as semi-structured interviews \cite{DBLP:journals/corr/abs-2305-13014}. These works reveal that LLMs perform well in handling different textual tasks and formats in thematic analysis.

\paragraph{Cultural and Historical Contexts Analysis by LLMs}

Investigations into the cultural and historical contexts within texts, particularly when analyzing with the aid of LLMs, demonstrate how well these models integrate and interpret such contexts. For instance, a benchmark for LLM-based machine translation shows that these models leverage historical and cultural information to improve translation accuracy and fidelity \cite{yao2024benchmarkingllmbasedmachinetranslation}. Researchers have explored the linguistic and stylistic patterns of LLM-generated texts, finding that while LLMs effectively produce text consistent with historical language styles, they still need refinement to fully capture the nuanced cultural tones of various literary traditions. Evaluations also suggest that despite improvements, identifying human oversight remains essential to ensure coherence in culturally rich translations \cite{karpinska2023documentleveltranslation}. Other studies reveal that LLMs can rank long-tail cultural concepts, indicating their ability to discern statistical trends in different regions \cite{jiang2023cpopqa}. Analysis based on Hofstede’s cultural dimensions shows that while LLMs align well with certain values, they struggle across diverse cultures \cite{masoud2023culturalalignment}.

\paragraph{Enhancing Reliability and Contextual Understanding in LLMs}

Efforts have also been made to create LLMs that are culturally aligned with specific linguistic communities. For example, models adapted for Traditional Chinese as used in Taiwan \cite{lin2023taiwanllmbridginglinguistic} and French \cite{faysse2024croissantllm} have been developed. These culturally resonant models demonstrate strong performance in understanding and generating text within their specific cultural contexts after the alignment.
Meanwhile, some studies have focused on enhancing the reliability of LLM outputs by embedding contextual information instead of creating a new LLM \cite{DBLP:journals/corr/abs-2408-04023}. Additionally, LLM-powered chatbots have been explored for their ability to dynamically provide responses sensitive to given culture. These chatbots are effective in accurately representing cultural nuances in heritage and cultural preservation contexts \cite{DBLP:conf/mmsys/RachabatuniPM024}. Moreover, research on LLMs has examined their capacity to adopt context-dependent value traits, proving their dynamic and controllable nature in contrast to the more stable traits observed in humans \cite{DBLP:journals/corr/abs-2307-07870}.

\paragraph{Moral and Psychological Dimensions of LLMs}

Regarding the moral and psychological aspects of LLMs \cite{ke2024exploringfrontiersllmspsychological,demszky2023using}, studies have investigated their reasoning about moral and legal issues \cite{ALMEIDA2024104145}. From a social psychology perspective, research on LLM agents reveals that these models can simulate and adapt collaborative behaviors based on contextual cues \cite{zhang2023exploring}. Additionally, efforts have been made to train psychologically specific models. For example, PsycoLLM was trained on a high-quality psychological dataset and has proven effective on psychology benchmarks \cite{hu2024psycollmenhancingllmpsychological}. LLM-powered agents also show significant potential in enhancing mental health support and serving as therapeutic tools \cite{lai2023psy,bill2023fine,wang2024towards}, demonstrating that LLMs can be effectively applied to psychological analysis.

\paragraph{LLMs in Literary Evaluation}

While previous studies have demonstrated various applications of LLMs, our work is pioneering in using LLMs as evaluators for analyzing Nobel Prize-winning literature. By leveraging the language understanding capabilities of LLMs and incorporating insights from engaged human evaluators, our study systematically assesses Nobel Prize-winning works across multiple dimensions. These dimensions include thematic analysis, intertextuality, cultural and historical contexts, linguistic and structural innovations, character development, moral and philosophical insights, narrative techniques, and emotional tone. Specifically, our novel approach not only quantifies literary merit but also evaluates how well LLM analyses align with human perspectives on literary aspects. By analyzing the gap between analysis from LLM and human interpretation, our study contributes to a more comprehensive understanding of literary merit in distinguished works of literature.

\section{Experiments and Observation }\label{sec3}
The central objective of this research is to evaluate the comparative capabilities of large language models (LLMs) and human participants in the execution of literary analysis, particularly focusing on works by Nobel Prize-winning authors. This investigation seeks to determine whether an advanced LLM, such as OpenAI's o1 model, can rival or surpass graduate-level students in interpreting and critiquing complex literary texts. Through the analysis of two short stories, "Nine Chapters" by Han Kang \cite{kang2016fruit} and "Friendship" by Jon Fosse \cite{fosse2018scenes}, this study endeavors to explore the extent to which AI models can engage with various dimensions of literary analysis, such as thematic interpretation, character development, and intertextuality, among others. 
Given the increasing application of AI in creative and intellectual fields, the significance of this research lies in its potential to illuminate the strengths and limitations of LLMs in tasks traditionally entrusted to human expertise. While o1 model has exhibited competence in structured data analysis and text generation, its efficacy in nuanced, subjective domains like literature remains insufficiently explored. This study thus contributes to the understanding of o1 model's role in the interpretation of humanistic content, offering insights that could influence the future integration of AI in literary studies and related disciplines. Additionally, by directly comparing the o1 model's output with that of expert human participants, this research establishes a structured framework for assessing how AI can complement or challenge conventional modes of literary critique, ultimately broadening the discourse on human-AI collaboration in the humanities.
\subsection{Experimental Design}\label{subsec2}
The design of this experiment follows a comparative framework, involving both human participants, who also contributed as coauthors of this paper, and the o1 model. They were given the task of analyzing two short stories that have won Nobel Prizes: "Nine Chapters" by Han Kang and "Friendship" by Jon Fosse. Participants, whether human or AI, conducted their analyses focusing on several literary factors: thematic exploration, intertextual connections, historical and cultural contexts, innovations in language and structure, character portrayal, moral and philosophical interpretations, narrative strategies, and the emotional and psychological depth. Here are the specific instructions of aspects to analyze:

\begin{itemize}
  \item \textbf{Thematic Analysis}: Identify the central themes of the work and discuss their significance in relation to the narrative’s broader context.
  \item \textbf{Intertextuality and Literary Influence}: Detect and comment on any references or influences from other literary traditions or works that shape the narrative's meaning.
  \item \textbf{Cultural and Historical Contexts}: Analyze how the story reflects or critiques the cultural and historical period in which it was written, offering insights into its societal implications.
  \item \textbf{Linguistic and Structural Innovations}: Highlight any innovative uses of language or narrative structures, such as non-linear storytelling or stylistic techniques like stream of consciousness.
  \item \textbf{Character Development}: Examine the development of central characters, focusing on their motivations, internal conflicts, and psychological depth.
  \item \textbf{Moral and Philosophical Insights}: Discuss any ethical dilemmas or philosophical themes present within the text and how these contribute to its deeper meaning.
  \item \textbf{Narrative Techniques and Temporal Structure}: Explore the narrative style, pacing, use of time, and any shifts in perspective that influence the reader's experience.
  \item \textbf{Emotional Tone and Psychological Depth}: Investigate the emotional atmosphere of the text, especially in how it conveys psychological complexity through the portrayal of characters.
\end{itemize}

These literary elements, which are provided as prompts to both human and LLM evaluators, form the core of evaluations, enabling structured comparisons across diverse interpretative aspects and ensuring analyses remain focused and aligned. For a systematic comparison, both qualitative and quantitative evaluation metrics were established. Qualitatively, human and AI responses were assessed for their capacity to capture intricate features of the texts, such as emotional cues, thematic sophistication, and philosophical reflections, offering a comprehensive comparison of engagement levels with the texts. 
Additionally, a quantitative assessment was conducted across three key dimensions: Coherence, Creativity, and Fidelity to the Text. Coherence examines the logical arrangement and clarity of the analysis and its effectiveness in constructing a convincing argument. Creativity evaluates the uniqueness of the insights provided, determining if the analysis introduces new viewpoints or simply reiterates conventional interpretations. Fidelity to the Text gauges how precisely the analysis aligns with the original content, ensuring interpretations are firmly rooted in the source material rather than conjecture.

Each of these quantitative dimensions was evaluated on a 1 to 5 scale, where 1 indicated minimal performance and 5 signified exceptional performance. The average scores for human and AI participants within each category were calculated, facilitating an objective and direct comparison. By integrating qualitative insights with quantitative evaluations, this study presents a comprehensive analysis of the relative strengths and weaknesses inherent in human versus AI literary analysis.

\subsection{Evaluation on Analysis of "Friendship"}\label{subsec3}

This section offers a thorough assessment of both the qualitative and quantitative dimensions of the literary analyses conducted by human participants and the o1 model. The purpose of this evaluation is to shed light on the strengths and weaknesses of each group when analyzing the Nobel Prize-winning short story "Friendship" by Jon Fosse.

\subsubsection{Qualitative Evaluation}

\paragraph{Coherence}
Human participants persistently exhibited a clear and logical organization in their responses, thereby sustaining flow and cohesion within their analyses. Their responses were frequently augmented with transitional phrases and logical connectors, facilitating a smooth progression of ideas. Conversely, the o1 model demonstrated certain deficiencies in coherence, particularly in discourse related to linguistic and structural innovations. The AI's outputs were insufficiently equipped with logical transitions, which at times resulted in fragmented or disjointed responses.

\paragraph{Creativity}
The o1 model outperformed human participants in certain creative aspects, particularly in its analysis of cultural background and intertextuality. It provided deeper and more detailed insights, such as drawing connections to existentialism and themes of alienation and the human condition. These responses revealed a broader and more nuanced understanding of the text. In contrast, while human participants produced more vivid and emotionally resonant analyses, their creative insights were not as expansive as those offered by the model in terms of connecting the text to broader literary traditions.

\paragraph{Fidelity to the Text}
In terms of fidelity, the o1 model excelled in producing highly accurate summaries of the text. It demonstrated a strong ability to remain closely aligned with the original content, particularly in character development and thematic analysis. For example, the model provided a thorough and precise summary of the main characters' traits, avoiding personal bias. Human participants, however, tended to introduce subjective interpretations, which sometimes deviated from the strict fidelity to the text. This subjective element was particularly noticeable in analyses concerning emotional tone and psychological depth, where human responses were influenced by personal perspectives.

\paragraph{Thematic Analysis}
Both the o1 model and human participants identified similar overarching themes, such as friendship, humanity, family, and society. However, the o1 model's thematic analysis was less refined in its language, lacking the aesthetic sophistication typically found in human-written responses. While the model presented a comprehensive view, human participants offered more nuanced, expressive interpretations of these themes.

\paragraph{Intertextuality and Literary Influence}
The o1 model provided deeper insights into intertextuality and literary influence, identifying significant literary connections, albeit with somewhat rigid expression. Human participants, by contrast, often drew from minimalist literary traditions, offering creative yet less technically accurate connections.

\paragraph{Cultural and Historical Contexts}
Both groups performed well in analyzing the cultural and historical context of the work, recognizing the relationship between the post-industrial society depicted and the individual. The o1 model was more precise in situating the story within its temporal and spatial context, while human participants offered broader, more interpretive insights.
\begin{figure}[H]
    \centering
    \includegraphics[width = .9\textwidth]{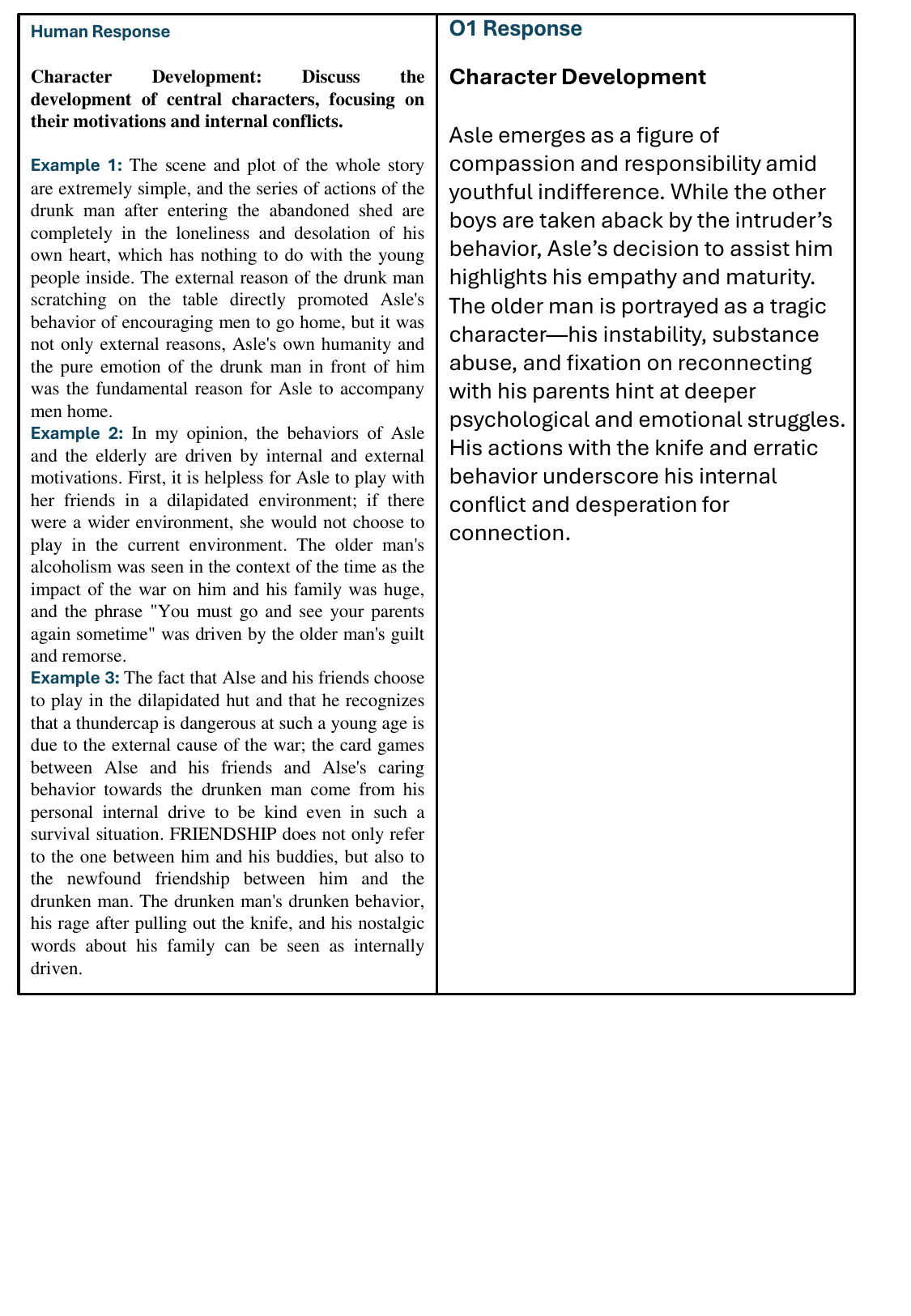}
    \caption[Character Development in Nobel Prize Literature]{\textbf{Character Development in Nobel Prize Literature "Friendship".} This figure illustrates the development of central characters in the Nobel Prize-winning literary work, comparing human analysis and the o1 model's response.}
    \label{fig:NOBEL_Character_Development}
\end{figure}
% \newpage
\paragraph{Linguistic and Structural Innovations}
The o1 model and human participants both highlighted the text's minimalist language and use of third-person perspective. However, human participants provided a more detailed analysis by noting specific symbols, such as "the red swastika," offering a more accurate interpretation of the text’s symbolic layers.

\paragraph{Character Development}
Both the o1 model and human participants analyzed the central characters effectively, focusing on the compassionate protagonist and the tragic elderly figure. Their analyses provided a well-rounded understanding of the characters' traits and development.

\paragraph{Moral and Philosophical Insights}
While both groups recognized social responsibility as a key theme, human participants offered a more comprehensive analysis by discussing the duality of violence and peace within the story. The o1 model, on the other hand, provided a more concise conclusion without delving into the reasoning behind its interpretation.

\paragraph{Narrative Techniques and Temporal Structure}
Both the o1 model and human participants noted the linear narrative structure of the story. However, the model provided more explicit examples to support its conclusions, offering a clearer explanation of the author's intent.

\paragraph{Emotional Tone and Psychological Depth}
The o1 model approached the emotional tone and psychological depth of the text by first summarizing key elements and then building upon them progressively. Human participants, on the other hand, used descriptive language and focused more on detailed environmental and character portrayals to convey their analyses.

Overall, the qualitative evaluation revealed that while the o1 model provided broader, more comprehensive analyses, it lacked the aesthetic and emotional nuance found in human interpretations. The model was objective and detailed, but human participants enriched their analyses with subjective insights, offering a more emotionally engaging experience.

\subsubsection{Quantitative Evaluation}

In addition to the qualitative assessment, the analysis was supplemented with quantitative evaluations across three key dimensions: coherence, creativity, and fidelity to the text. Each dimension was rated on a scale of 1 to 5, with 1 representing minimal performance and 5 representing exceptional performance. The average scores for human participants and the o1 model are summarized in the table below:

\begin{table}[h]
\centering
\begin{tabular}{|l|c|c|c|}
\hline
\textbf{Name} & \textbf{Coherence} & \textbf{Creativity} & \textbf{Fidelity to the Text} \\ \hline
\textbf{Human Mean} & 4.19 & 4.19 & 4.16 \\ \hline
\textbf{o1 Model} & 3.00 & 4.25 & 4.50 \\ \hline
\end{tabular}
\caption{Quantitative Evaluation Results for Analysis of ”Friendship”.}
\end{table}

The results indicate that human participants outperformed the o1 model in terms of coherence, with an average score of 4.19 compared to the model's 3.00. This aligns with the qualitative findings, where human participants exhibited more logically structured and cohesive responses.

However, in the dimension of creativity, the o1 model slightly surpassed the human mean, achieving a score of 4.25 compared to 4.19. The model's ability to provide innovative insights into cultural contexts and intertextuality contributed to this higher score.

Finally, the o1 model outperformed human participants in fidelity to the text, scoring 4.50 compared to the human mean of 4.16. The model’s ability to maintain close adherence to the original text, particularly in character and thematic analysis, accounts for this advantage.

\subsection{Evaluation on Analysis of "Nine Chapters"}

\subsubsection{Qualitative Analysis}

The qualitative evaluation focuses on comparing the literary analysis results produced by human participants and the o1 model on Han Kang's work \textit{The Vegetarian}, specifically the chapter titled "Nine Chapters." The analysis explores various literary aspects, including thematic analysis, coherence, creativity, and fidelity to the text.

\paragraph{Thematic Analysis}
Both the human participants and the o1 model identified key themes such as love, freedom, and solitude. The o1 model demonstrated an innovative approach by incorporating time progression, aging, nature, and transcendence as additional thematic layers, though its language was at times rigid, making comprehension challenging. In contrast, some human participants introduced themes more directly tied to the content, such as social and gender relations, which offered more contextually relevant insights.

\paragraph{Coherence}
Human participants generally presented their ideas with clarity, utilizing logical connectors such as "on the one hand" or "in conclusion," which contributed to the flow of their analysis. The o1 model, by contrast, often presented its findings in a list-like manner, reducing the overall coherence and logical development of its arguments.

\paragraph{Creativity}
The o1 model showcased notable creativity in its analysis, particularly in the areas of cultural background, intertextuality, literary influence, and linguistic innovation. It accurately highlighted the socio-cultural implications of urbanization and even pinpointed the late 20th-century as a significant historical context. In intertextual analysis, the o1 model demonstrated a deeper understanding by referencing modernist writers. The model also provided comprehensive insights into narrative techniques, discussing narrative perspective, pacing, and dream sequences more thoroughly than the human participants, who generally confined their analysis to narrative perspective alone.

\paragraph{Fidelity to the Text}
The o1 model received high scores for fidelity to the text due to its reliance on the material directly from the narrative. However, human participants tended to incorporate broader knowledge, including societal pressures and gender dynamics relevant to the entire book, leading to a more nuanced but slightly less textually faithful interpretation. For example, while the o1 model stuck closely to the text's emotional tone and psychological depth, human participants offered deeper reflections on existential themes such as self-worth and social justice for women, albeit at the cost of strict adherence to the text.

\paragraph{Intertextuality and Literary Influence}
In this category, the o1 model offered a more accurate and thorough assessment. It drew connections between the novel and sources of inspiration, such as Chinese classical poetry and modern literary traditions, and employed well-chosen examples to support its claims. Human participants tended to focus on a single point of analysis, with only a few recognizing the poetic qualities of the text.

\paragraph{Cultural and Historical Background}
Both the o1 model and human participants recognized the theme of urbanization and societal pressures present in the work. However, the o1 model's discussion of social expectations was somewhat disjointed from the text, while human participants offered a more relevant perspective by emphasizing the role of women's social status in the novel's context.

\paragraph{Linguistic Innovation}
Both human and o1 model analyses identified the minimalist and modern narrative techniques employed by the author. The o1 model offered concise explanations of literary terms, while human participants provided more detailed examples directly from the text, offering a more grounded and logical perspective.

\paragraph{Character Development}
The o1 model's concise language contrasted with the more detailed examples provided by human participants, who expanded on character traits with extensive textual references. The o1 model's analysis resembled a standard academic response, while the human responses provided richer, more descriptive insights.
\begin{figure}[H]
    \centering
    \includegraphics[width = .9\textwidth]{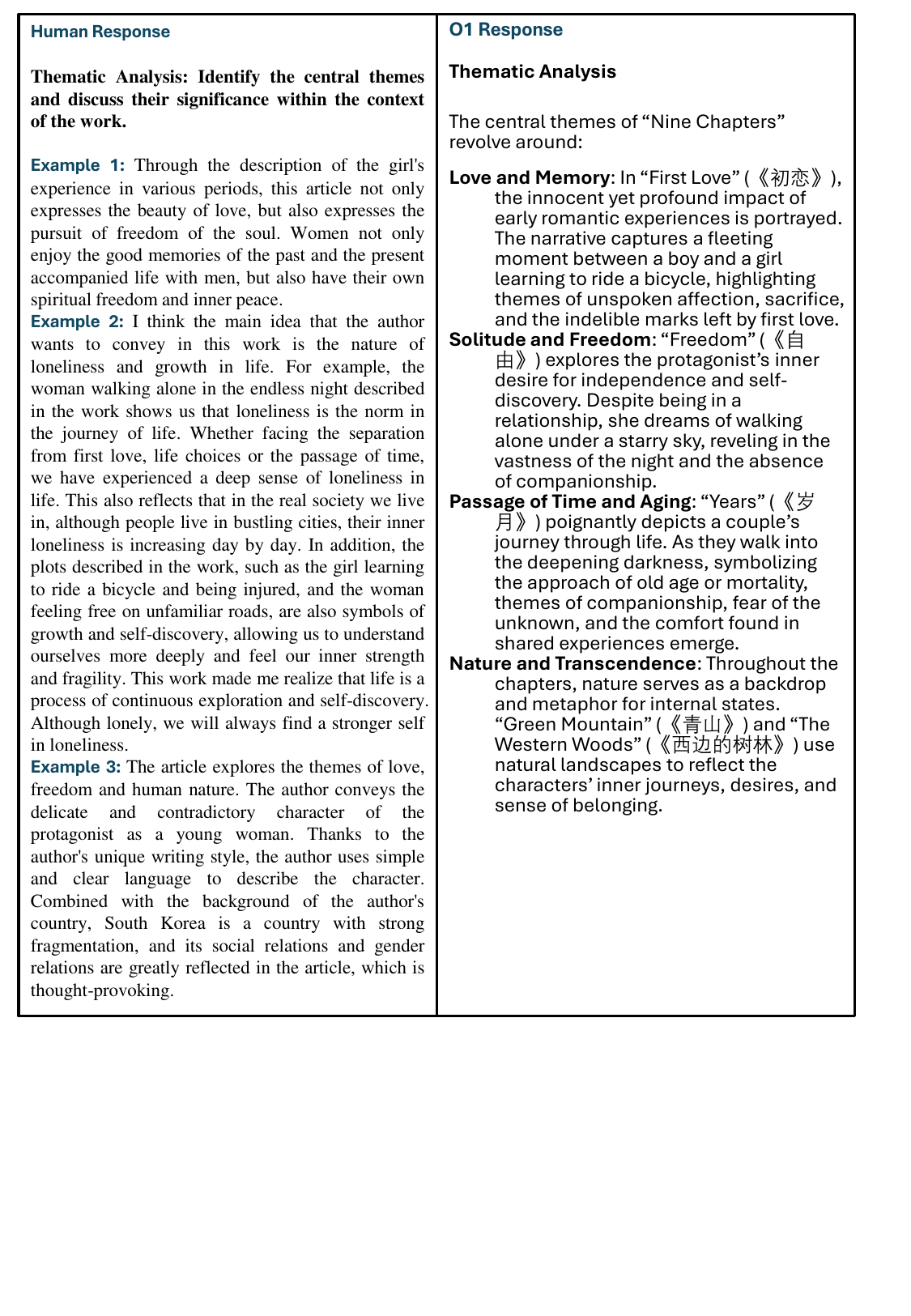}
    \caption[Thematic Analysis in Nobel Prize Literature]{\textbf{Thematic Analysis in Nobel Prize Literature "Nine Chapters".} This figure presents an analysis of the central themes in the Nobel Prize-winning literary work, comparing human analysis and the o1 model's response.}
\end{figure}
\newpage
\paragraph{Moral and Philosophical Insights}
Both the o1 model and human participants accurately identified core themes, such as gender dynamics and societal expectations. The o1 model employed technical terminology to summarize these themes, while human participants displayed more originality in their reflections, especially concerning the philosophical contemplation of gender roles in society.

\paragraph{Narrative Techniques and Temporal Structure}
The o1 model offered a broader analysis of the narrative structure, including detailed discussion of narrative perspective and pacing. In contrast, human participants often limited their analysis to a specific narrative viewpoint but provided a more engaging, vivid description, aligning more closely with the original text.

\paragraph{Emotional Tone and Psychological Depth}
Human participants used descriptive language to analyze emotional tone and psychological depth, often linking these to specific textual elements, which enhanced their interpretative depth. The o1 model's analysis, while accurate, lacked the detailed, layered interpretation provided by the human participants.

\subsubsection{Quantitative Analysis}

The quantitative results focus on three key dimensions of literary analysis: coherence, creativity, and fidelity to the text. The following table summarizes the mean scores from human participants and the o1 model across these categories.

\begin{table}[h!]
\centering
\begin{tabular}{|l|c|c|c|}
\hline
\textbf{Name}      & \textbf{Coherence} & \textbf{Creativity} & \textbf{Fidelity to the Text} \\ \hline
\textbf{human mean} & 4.31875            & 4.0625              & 4.24375                      \\ \hline
\textbf{o1 model}          & 4.0                & 4.2                 & 4.3                          \\ \hline
\end{tabular}
\caption{Quantitative Evaluation Results for Analysis of ”Nine Chapters”}
\end{table}

The table indicates that human participants slightly outperformed the o1 model in terms of coherence, likely due to their more fluid and logically connected responses. However, the o1 model demonstrated a marginal advantage in creativity and fidelity to the text, reflecting its ability to provide more detailed cultural and intertextual analysis as well as its strict adherence to the textual content.

\section{Discussion}
The results of our comparative study between human participants and the o1 model in literary analysis reveal both the strengths and limitations of LLMs in interpreting complex texts such as Nobel Prize-winning literature. This research provides valuable insights into the growing capabilities of LLMs in domains traditionally dominated by human experts.

\paragraph{LLMs as Competent Literary Analysts}

The quantitative and qualitative evaluations demonstrate that the o1 model performs comparably to human participants in several dimensions, particularly in creativity and fidelity to the text. The model’s ability to generate innovative connections within the literary framework, such as identifying intertextuality and cultural contexts, showcases its potential to contribute meaningfully to literary analysis. In particular, the o1 model excelled in providing detailed, precise thematic and character analyses while maintaining a high degree of fidelity to the source material. These findings indicate that LLMs can offer objective, text-based interpretations that may complement or enhance human expertise, especially in tasks requiring meticulous content adherence.

\paragraph{Challenges in Aesthetic and Emotional Nuance}

Despite its strong performance in structured, text-based evaluation, the o1 model exhibited notable deficiencies in coherence and emotional depth. Human participants consistently outperformed the model in these areas, reflecting their superior ability to connect thematic elements with nuanced emotional and psychological insights. The human-generated analyses were often more coherent and fluid, demonstrating a deeper engagement with the subtleties of narrative tone and character development. These results highlight the continued importance of human interpretation in tasks that require a high degree of subjectivity and emotional resonance, aspects that current LLMs are not yet fully equipped to replicate.

\paragraph{Implications for Human-AI Collaboration in Literary Studies}

Our findings suggest that while LLMs like the o1 model are capable of performing rigorous textual analysis, they are best suited to roles where objective accuracy and creativity are prioritized over emotional and aesthetic interpretations. This opens opportunities for human-AI collaboration in the humanities, where LLMs can provide preliminary analyses, identify intertextual and cultural patterns, and perform detailed thematic evaluations, allowing human experts to focus on more interpretative and emotional aspects of literary critique.

\paragraph{Future Directions}
Recent studies highlight how the rapid advancement of AGI is transforming the arts and humanities, with LLMs showing impressive capabilities across various artistic fields, including text analysis ranging of poetry, history, and beyond. However, this raises critical concerns about truth, bias, accountability, and social impact~\cite{liu2023transformation}. While this study adds to the growing literature on AI in the humanities, demonstrating both the current strengths of LLMs and the areas where human expertise remains indispensable, there are still several key aresa for improvemwnt: 

First, enhancing the coherence and emotional interpretive abilities of LLMs is essential. Future models could be designed to incorporate elements of human emotional reasoning, which would help in bridging the gap between mechanical text analysis and humanistic literary critique. 
Second, extending this study to larger datasets and more diverse literary genres could provide a broader understanding of LLMs' capabilities and limitations in different contexts. Finally, technical solutions such as robust factuality evaluations, toxicity filters, and bias detectors will be vital to ensureing LLMs' reliability and trustworthiness. These will help ensure the uphold cultural values, pluralism, and dignity.

Overall, this study contributes to the growing body of literature on AI in the humanities, demonstrating both the current strengths of LLMs and the areas where human expertise remains indispensable. As AI continues to evolve, the collaboration between human and machine in creative fields will likely lead to more sophisticated and insightful analyses, furthering our understanding of both literature and AI's role in intellectual discourse.

\bibliography{sn-bibliography}% common bib file


\begin{thebibliography}{40}
\ifx \bisbn   \undefined \def \bisbn  #1{ISBN #1}\fi
\ifx \binits  \undefined \def \binits#1{#1}\fi
\ifx \bauthor  \undefined \def \bauthor#1{#1}\fi
\ifx \batitle  \undefined \def \batitle#1{#1}\fi
\ifx \bjtitle  \undefined \def \bjtitle#1{#1}\fi
\ifx \bvolume  \undefined \def \bvolume#1{\textbf{#1}}\fi
\ifx \byear  \undefined \def \byear#1{#1}\fi
\ifx \bissue  \undefined \def \bissue#1{#1}\fi
\ifx \bfpage  \undefined \def \bfpage#1{#1}\fi
\ifx \blpage  \undefined \def \blpage #1{#1}\fi
\ifx \burl  \undefined \def \burl#1{\textsf{#1}}\fi
\ifx \doiurl  \undefined \def \doiurl#1{\url{https://doi.org/#1}}\fi
\ifx \betal  \undefined \def \betal{\textit{et al.}}\fi
\ifx \binstitute  \undefined \def \binstitute#1{#1}\fi
\ifx \binstitutionaled  \undefined \def \binstitutionaled#1{#1}\fi
\ifx \bctitle  \undefined \def \bctitle#1{#1}\fi
\ifx \beditor  \undefined \def \beditor#1{#1}\fi
\ifx \bpublisher  \undefined \def \bpublisher#1{#1}\fi
\ifx \bbtitle  \undefined \def \bbtitle#1{#1}\fi
\ifx \bedition  \undefined \def \bedition#1{#1}\fi
\ifx \bseriesno  \undefined \def \bseriesno#1{#1}\fi
\ifx \blocation  \undefined \def \blocation#1{#1}\fi
\ifx \bsertitle  \undefined \def \bsertitle#1{#1}\fi
\ifx \bsnm \undefined \def \bsnm#1{#1}\fi
\ifx \bsuffix \undefined \def \bsuffix#1{#1}\fi
\ifx \bparticle \undefined \def \bparticle#1{#1}\fi
\ifx \barticle \undefined \def \barticle#1{#1}\fi
\bibcommenthead
\ifx \bconfdate \undefined \def \bconfdate #1{#1}\fi
\ifx \botherref \undefined \def \botherref #1{#1}\fi
\ifx \url \undefined \def \url#1{\textsf{#1}}\fi
\ifx \bchapter \undefined \def \bchapter#1{#1}\fi
\ifx \bbook \undefined \def \bbook#1{#1}\fi
\ifx \bcomment \undefined \def \bcomment#1{#1}\fi
\ifx \oauthor \undefined \def \oauthor#1{#1}\fi
\ifx \citeauthoryear \undefined \def \citeauthoryear#1{#1}\fi
\ifx \endbibitem  \undefined \def \endbibitem {}\fi
\ifx \bconflocation  \undefined \def \bconflocation#1{#1}\fi
\ifx \arxivurl  \undefined \def \arxivurl#1{\textsf{#1}}\fi
\csname PreBibitemsHook\endcsname

%%% 1
\bibitem[\protect\citeauthoryear{Cardoso et~al.}{2009}]{cardoso2009converging}
\begin{barticle}
\bauthor{\bsnm{Cardoso}, \binits{A.}},
\bauthor{\bsnm{Veale}, \binits{T.}},
\bauthor{\bsnm{Wiggins}, \binits{G.A.}}:
\batitle{Converging on the divergent: The history (and future) of the international joint workshops in computational creativity}.
\bjtitle{AI magazine}
\bvolume{30}(\bissue{3}),
\bfpage{15}--\blpage{15}
(\byear{2009})
\end{barticle}
\endbibitem

%%% 2
\bibitem[\protect\citeauthoryear{Shanahan and Clarke}{2023}]{shanahan2023evaluating}
\begin{botherref}
\oauthor{\bsnm{Shanahan}, \binits{M.}},
\oauthor{\bsnm{Clarke}, \binits{C.}}:
Evaluating large language model creativity from a literary perspective.
arXiv preprint arXiv:2312.03746
(2023)
\end{botherref}
\endbibitem

%%% 3
\bibitem[\protect\citeauthoryear{OpenAI et~al.}{2024}]{openai2024gpt}
\begin{botherref}
\oauthor{\bsnm{OpenAI}, \binits{A.J.}},
\oauthor{\bsnm{Adler}, \binits{S.}},
\oauthor{\bsnm{Agarwal}, \binits{S.}},
\oauthor{\bsnm{Ahmad}, \binits{L.}},
\oauthor{\bsnm{Akkaya}, \binits{I.}},
\oauthor{\bsnm{Aleman}, \binits{F.L.}},
\oauthor{\bsnm{Almeida}, \binits{D.}},
\oauthor{\bsnm{Altenschmidt}, \binits{J.}},
\oauthor{\bsnm{Altman}, \binits{S.}},
\oauthor{\bsnm{Anadkat}, \binits{S.}}, et al.:
Gpt-4 technical report, 2024.
URL https://arxiv. org/abs/2303.08774
(2024)
\end{botherref}
\endbibitem

%%% 4
\bibitem[\protect\citeauthoryear{Team et~al.}{2023}]{team2023gemini}
\begin{botherref}
\oauthor{\bsnm{Team}, \binits{G.}},
\oauthor{\bsnm{Anil}, \binits{R.}},
\oauthor{\bsnm{Borgeaud}, \binits{S.}},
\oauthor{\bsnm{Wu}, \binits{Y.}},
\oauthor{\bsnm{Alayrac}, \binits{J.-B.}},
\oauthor{\bsnm{Yu}, \binits{J.}},
\oauthor{\bsnm{Soricut}, \binits{R.}},
\oauthor{\bsnm{Schalkwyk}, \binits{J.}},
\oauthor{\bsnm{Dai}, \binits{A.M.}},
\oauthor{\bsnm{Hauth}, \binits{A.}}, et al.:
Gemini: a family of highly capable multimodal models.
arXiv preprint arXiv:2312.11805
(2023)
\end{botherref}
\endbibitem

%%% 5
\bibitem[\protect\citeauthoryear{Touvron et~al.}{2023}]{touvron2023llama}
\begin{botherref}
\oauthor{\bsnm{Touvron}, \binits{H.}},
\oauthor{\bsnm{Lavril}, \binits{T.}},
\oauthor{\bsnm{Izacard}, \binits{G.}},
\oauthor{\bsnm{Martinet}, \binits{X.}},
\oauthor{\bsnm{Lachaux}, \binits{M.-A.}},
\oauthor{\bsnm{Lacroix}, \binits{T.}},
\oauthor{\bsnm{Rozi{\`e}re}, \binits{B.}},
\oauthor{\bsnm{Goyal}, \binits{N.}},
\oauthor{\bsnm{Hambro}, \binits{E.}},
\oauthor{\bsnm{Azhar}, \binits{F.}}, et al.:
Llama: Open and efficient foundation language models.
arXiv preprint arXiv:2302.13971
(2023)
\end{botherref}
\endbibitem

%%% 6
\bibitem[\protect\citeauthoryear{Zhong et~al.}{2024}]{zhong2024evaluation}
\begin{botherref}
\oauthor{\bsnm{Zhong}, \binits{T.}},
\oauthor{\bsnm{Liu}, \binits{Z.}},
\oauthor{\bsnm{Pan}, \binits{Y.}},
\oauthor{\bsnm{Zhang}, \binits{Y.}},
\oauthor{\bsnm{Zhou}, \binits{Y.}},
\oauthor{\bsnm{Liang}, \binits{S.}},
\oauthor{\bsnm{Wu}, \binits{Z.}},
\oauthor{\bsnm{Lyu}, \binits{Y.}},
\oauthor{\bsnm{Shu}, \binits{P.}},
\oauthor{\bsnm{Yu}, \binits{X.}}, et al.:
Evaluation of openai o1: Opportunities and challenges of agi.
arXiv preprint arXiv:2409.18486
(2024)
\end{botherref}
\endbibitem

%%% 7
\bibitem[\protect\citeauthoryear{Zhao et~al.}{2024}]{zhao2024understanding}
\begin{botherref}
\oauthor{\bsnm{Zhao}, \binits{C.}},
\oauthor{\bsnm{Wang}, \binits{B.}},
\oauthor{\bsnm{Wang}, \binits{Z.}}:
Understanding literary texts by llms: A case study of ancient chinese poetry.
arXiv preprint arXiv:2409.00060
(2024)
\end{botherref}
\endbibitem

%%% 8
\bibitem[\protect\citeauthoryear{Nikolova-Stoupak et~al.}{2024}]{nikolova2024contemporary}
\begin{bchapter}
\bauthor{\bsnm{Nikolova-Stoupak}, \binits{I.}},
\bauthor{\bsnm{Lejeune}, \binits{G.}},
\bauthor{\bsnm{Schaeffer-Lacroix}, \binits{E.}}:
\bctitle{Contemporary llms and literary abridgement: An analytical inquiry}.
In: \bbtitle{Sixth International Conference},
p. \bfpage{39}
(\byear{2024})
\end{bchapter}
\endbibitem

%%% 9
\bibitem[\protect\citeauthoryear{Michel et~al.}{2024}]{michel2024realistic}
\begin{botherref}
\oauthor{\bsnm{Michel}, \binits{G.}},
\oauthor{\bsnm{Epure}, \binits{E.V.}},
\oauthor{\bsnm{Hennequin}, \binits{R.}},
\oauthor{\bsnm{Cerisara}, \binits{C.}}:
A realistic evaluation of llms for quotation attribution in literary texts: A case study of llama3.
arXiv preprint arXiv:2406.11380
(2024)
\end{botherref}
\endbibitem

%%% 10
\bibitem[\protect\citeauthoryear{Yu et~al.}{2024}]{yu2024lfed}
\begin{botherref}
\oauthor{\bsnm{Yu}, \binits{L.}},
\oauthor{\bsnm{Liu}, \binits{Q.}},
\oauthor{\bsnm{Xiong}, \binits{D.}}:
Lfed: A literary fiction evaluation dataset for large language models.
arXiv preprint arXiv:2405.10166
(2024)
\end{botherref}
\endbibitem

%%% 11
\bibitem[\protect\citeauthoryear{Zhang et~al.}{2024}]{zhang2024binarized}
\begin{botherref}
\oauthor{\bsnm{Zhang}, \binits{Y.}},
\oauthor{\bsnm{Garg}, \binits{A.}},
\oauthor{\bsnm{Cao}, \binits{Y.}},
\oauthor{\bsnm{Lew}, \binits{L.}},
\oauthor{\bsnm{Ghorbani}, \binits{B.}},
\oauthor{\bsnm{Zhang}, \binits{Z.}},
\oauthor{\bsnm{Firat}, \binits{O.}}:
Binarized neural machine translation.
Advances in Neural Information Processing Systems
\textbf{36}
(2024)
\end{botherref}
\endbibitem

%%% 12
\bibitem[\protect\citeauthoryear{Li et~al.}{2024}]{li2024pre}
\begin{barticle}
\bauthor{\bsnm{Li}, \binits{J.}},
\bauthor{\bsnm{Tang}, \binits{T.}},
\bauthor{\bsnm{Zhao}, \binits{W.X.}},
\bauthor{\bsnm{Nie}, \binits{J.-Y.}},
\bauthor{\bsnm{Wen}, \binits{J.-R.}}:
\batitle{Pre-trained language models for text generation: A survey}.
\bjtitle{ACM Computing Surveys}
\bvolume{56}(\bissue{9}),
\bfpage{1}--\blpage{39}
(\byear{2024})
\end{barticle}
\endbibitem

%%% 13
\bibitem[\protect\citeauthoryear{Kaddour et~al.}{2023}]{Kaddour2023Challenges}
\begin{botherref}
\oauthor{\bsnm{Kaddour}, \binits{J.}},
\oauthor{\bsnm{Harris}, \binits{J.}},
\oauthor{\bsnm{Mozes}, \binits{M.}},
\oauthor{\bsnm{Bradley}, \binits{H.}},
\oauthor{\bsnm{Raileanu}, \binits{R.}},
\oauthor{\bsnm{McHardy}, \binits{R.}}:
Challenges and applications of large language models.
ArXiv
\textbf{abs/2307.10169}
(2023)
\doiurl{10.48550/arXiv.2307.10169}
\end{botherref}
\endbibitem

%%% 14
\bibitem[\protect\citeauthoryear{Wang et~al.}{2023}]{Wang2023Document-Level}
\begin{botherref}
\oauthor{\bsnm{Wang}, \binits{L.}},
\oauthor{\bsnm{Lyu}, \binits{C.}},
\oauthor{\bsnm{Ji}, \binits{T.}},
\oauthor{\bsnm{Zhang}, \binits{Z.}},
\oauthor{\bsnm{Yu}, \binits{D.}},
\oauthor{\bsnm{Shi}, \binits{S.}},
\oauthor{\bsnm{Tu}, \binits{Z.}}:
Document-level machine translation with large language models.
ArXiv
\textbf{abs/2304.02210}
(2023)
\doiurl{10.48550/arXiv.2304.02210}
\end{botherref}
\endbibitem

%%% 15
\bibitem[\protect\citeauthoryear{Chakrabarty et~al.}{2023}]{Chakrabarty2023Art}
\begin{botherref}
\oauthor{\bsnm{Chakrabarty}, \binits{T.}},
\oauthor{\bsnm{Laban}, \binits{P.}},
\oauthor{\bsnm{Agarwal}, \binits{D.}},
\oauthor{\bsnm{Muresan}, \binits{S.}},
\oauthor{\bsnm{Wu}, \binits{C.-S.}}:
Art or artifice? large language models and the false promise of creativity.
ArXiv
\textbf{abs/2309.14556}
(2023)
\doiurl{10.48550/arXiv.2309.14556}
\end{botherref}
\endbibitem

%%% 16
\bibitem[\protect\citeauthoryear{Dai et~al.}{2023}]{DBLP:conf/emnlp/DaiXK}
\begin{bchapter}
\bauthor{\bsnm{Dai}, \binits{S.}},
\bauthor{\bsnm{Xiong}, \binits{A.}},
\bauthor{\bsnm{Ku}, \binits{L.}}:
\bctitle{Llm-in-the-loop: Leveraging large language model for thematic analysis}.
In: \beditor{\bsnm{Bouamor}, \binits{H.}},
\beditor{\bsnm{Pino}, \binits{J.}},
\beditor{\bsnm{Bali}, \binits{K.}} (eds.)
\bbtitle{Findings of the Association for Computational Linguistics: {EMNLP} 2023, Singapore, December 6-10, 2023},
pp. \bfpage{9993}--\blpage{10001}
(\byear{2023}).
\doiurl{10.18653/V1/2023.FINDINGS-EMNLP.669}
\end{bchapter}
\endbibitem

%%% 17
\bibitem[\protect\citeauthoryear{Paoli}{2024}]{DBLP:journals/corr/abs-2404-08488}
\begin{botherref}
\oauthor{\bsnm{Paoli}, \binits{S.D.}}:
Thematic analysis with large language models: does it work with languages other than english? {A} targeted test in italian.
CoRR
\textbf{abs/2404.08488}
(2024)
\doiurl{10.48550/ARXIV.2404.08488}
{\href{https://arxiv.org/abs/2404.08488}{{2404.08488}}}
\end{botherref}
\endbibitem

%%% 18
\bibitem[\protect\citeauthoryear{Dr{\'{a}}pal et~al.}{2023}]{DBLP:conf/jurix/DrapalWS23}
\begin{bchapter}
\bauthor{\bsnm{Dr{\'{a}}pal}, \binits{J.}},
\bauthor{\bsnm{Westermann}, \binits{H.}},
\bauthor{\bsnm{Savelka}, \binits{J.}}:
\bctitle{Using large language models to support thematic analysis in empirical legal studies}.
In: \beditor{\bsnm{Sileno}, \binits{G.}},
\beditor{\bsnm{Spanakis}, \binits{J.}},
\beditor{\bsnm{Dijck}, \binits{G.}} (eds.)
\bbtitle{Legal Knowledge and Information Systems - {JURIX} 2023: The Thirty-sixth Annual Conference, Maastricht, The Netherlands, 18-20 December 2023}.
\bsertitle{Frontiers in Artificial Intelligence and Applications},
vol. \bseriesno{379},
pp. \bfpage{197}--\blpage{206}
(\byear{2023}).
\doiurl{10.3233/FAIA230965}
\end{bchapter}
\endbibitem

%%% 19
\bibitem[\protect\citeauthoryear{Choi}{2023}]{Poetry}
\begin{barticle}
\bauthor{\bsnm{Choi}, \binits{K.}}:
\batitle{Computational thematic analysis of poetry via bimodal large language models}.
\bjtitle{Proceedings of the Association for Information Science and Technology}
\bvolume{60}(\bissue{1}),
\bfpage{538}--\blpage{542}
(\byear{2023})
\doiurl{10.1002/pra2.812}
{\href{https://arxiv.org/abs/https://asistdl.onlinelibrary.wiley.com/doi/pdf/10.1002/pra2.812}{{https://asistdl.onlinelibrary.wiley.com/doi/pdf/10.1002/pra2.812}}}
\end{barticle}
\endbibitem

%%% 20
\bibitem[\protect\citeauthoryear{Paoli}{2023}]{DBLP:journals/corr/abs-2305-13014}
\begin{botherref}
\oauthor{\bsnm{Paoli}, \binits{S.D.}}:
Can large language models emulate an inductive thematic analysis of semi-structured interviews? an exploration and provocation on the limits of the approach and the model.
CoRR
\textbf{abs/2305.13014}
(2023)
\doiurl{10.48550/ARXIV.2305.13014}
{\href{https://arxiv.org/abs/2305.13014}{{2305.13014}}}
\end{botherref}
\endbibitem

%%% 21
\bibitem[\protect\citeauthoryear{Yao et~al.}{2024}]{yao2024benchmarkingllmbasedmachinetranslation}
\begin{botherref}
\oauthor{\bsnm{Yao}, \binits{B.}},
\oauthor{\bsnm{Jiang}, \binits{M.}},
\oauthor{\bsnm{Yang}, \binits{D.}},
\oauthor{\bsnm{Hu}, \binits{J.}}:
Benchmarking LLM-based Machine Translation on Cultural Awareness
(2024).
\url{https://arxiv.org/abs/2305.14328}
\end{botherref}
\endbibitem

%%% 22
\bibitem[\protect\citeauthoryear{Karpinska and Iyyer}{2023}]{karpinska2023documentleveltranslation}
\begin{botherref}
\oauthor{\bsnm{Karpinska}, \binits{M.}},
\oauthor{\bsnm{Iyyer}, \binits{M.}}:
Large language models effectively leverage document-level context for literary translation, but critical errors persist.
ArXiv
\textbf{abs/2304.03245}
(2023)
\doiurl{10.48550/arXiv.2304.03245}
\end{botherref}
\endbibitem

%%% 23
\bibitem[\protect\citeauthoryear{Jiang and Joshi}{2023}]{jiang2023cpopqa}
\begin{botherref}
\oauthor{\bsnm{Jiang}, \binits{M.}},
\oauthor{\bsnm{Joshi}, \binits{M.}}:
Cpopqa: Ranking cultural concept popularity by llms.
ArXiv
\textbf{abs/2311.07897}
(2023)
\doiurl{10.48550/arXiv.2311.07897}
\end{botherref}
\endbibitem

%%% 24
\bibitem[\protect\citeauthoryear{Masoud et~al.}{2023}]{masoud2023culturalalignment}
\begin{botherref}
\oauthor{\bsnm{Masoud}, \binits{R.I.}},
\oauthor{\bsnm{Liu}, \binits{Z.}},
\oauthor{\bsnm{Ferianc}, \binits{M.}},
\oauthor{\bsnm{Treleaven}, \binits{P.C.}},
\oauthor{\bsnm{Rodrigues}, \binits{M.}}:
Cultural alignment in large language models: An explanatory analysis based on hofstede's cultural dimensions.
ArXiv
\textbf{abs/2309.12342}
(2023)
\doiurl{10.48550/arXiv.2309.12342}
\end{botherref}
\endbibitem

%%% 25
\bibitem[\protect\citeauthoryear{Lin and Chen}{2023}]{lin2023taiwanllmbridginglinguistic}
\begin{botherref}
\oauthor{\bsnm{Lin}, \binits{Y.-T.}},
\oauthor{\bsnm{Chen}, \binits{Y.-N.}}:
Taiwan LLM: Bridging the Linguistic Divide with a Culturally Aligned Language Model
(2023).
\url{https://arxiv.org/abs/2311.17487}
\end{botherref}
\endbibitem

%%% 26
\bibitem[\protect\citeauthoryear{Faysse et~al.}{2024}]{faysse2024croissantllm}
\begin{botherref}
\oauthor{\bsnm{Faysse}, \binits{M.}},
\oauthor{\bsnm{Fernandes}, \binits{P.}},
\oauthor{\bsnm{Guerreiro}, \binits{N.M.}},
\oauthor{\bsnm{Loison}, \binits{A.}},
\oauthor{\bsnm{Alves}, \binits{D.M.}},
\oauthor{\bsnm{Corro}, \binits{C.}},
\oauthor{\bsnm{Boizard}, \binits{N.}},
\oauthor{\bsnm{Alves}, \binits{J.}},
\oauthor{\bsnm{Rei}, \binits{R.}},
\oauthor{\bsnm{Martins}, \binits{P.H.}},
\oauthor{\bsnm{Casademunt}, \binits{A.B.}},
\oauthor{\bsnm{Yvon}, \binits{F.}},
\oauthor{\bsnm{Martins}, \binits{A.F.T.}},
\oauthor{\bsnm{Viaud}, \binits{G.}},
\oauthor{\bsnm{Hudelot}, \binits{C.}},
\oauthor{\bsnm{Colombo}, \binits{P.}}:
CroissantLLM: A Truly Bilingual French-English Language Model
(2024).
\url{https://arxiv.org/abs/2402.00786}
\end{botherref}
\endbibitem

%%% 27
\bibitem[\protect\citeauthoryear{Talukdar and Biswas}{2024}]{DBLP:journals/corr/abs-2408-04023}
\begin{botherref}
\oauthor{\bsnm{Talukdar}, \binits{W.}},
\oauthor{\bsnm{Biswas}, \binits{A.}}:
Improving large language model {(LLM)} fidelity through context-aware grounding: {A} systematic approach to reliability and veracity.
CoRR
\textbf{abs/2408.04023}
(2024)
\doiurl{10.48550/ARXIV.2408.04023}
{\href{https://arxiv.org/abs/2408.04023}{{2408.04023}}}
\end{botherref}
\endbibitem

%%% 28
\bibitem[\protect\citeauthoryear{Rachabatuni et~al.}{2024}]{DBLP:conf/mmsys/RachabatuniPM024}
\begin{bchapter}
\bauthor{\bsnm{Rachabatuni}, \binits{P.K.}},
\bauthor{\bsnm{Principi}, \binits{F.}},
\bauthor{\bsnm{Mazzanti}, \binits{P.}},
\bauthor{\bsnm{Bertini}, \binits{M.}}:
\bctitle{Context-aware chatbot using mllms for cultural heritage}.
In: \bbtitle{Proceedings of the 15th {ACM} Multimedia Systems Conference, MMSys 2024, Bari, Italy, April 15-18, 2024},
pp. \bfpage{459}--\blpage{463}
(\byear{2024}).
\doiurl{10.1145/3625468.3652193}
\end{bchapter}
\endbibitem

%%% 29
\bibitem[\protect\citeauthoryear{Kovac et~al.}{2023}]{DBLP:journals/corr/abs-2307-07870}
\begin{botherref}
\oauthor{\bsnm{Kovac}, \binits{G.}},
\oauthor{\bsnm{Sawayama}, \binits{M.}},
\oauthor{\bsnm{Portelas}, \binits{R.}},
\oauthor{\bsnm{Colas}, \binits{C.}},
\oauthor{\bsnm{Dominey}, \binits{P.F.}},
\oauthor{\bsnm{Oudeyer}, \binits{P.}}:
Large language models as superpositions of cultural perspectives.
CoRR
\textbf{abs/2307.07870}
(2023)
\doiurl{10.48550/ARXIV.2307.07870}
{\href{https://arxiv.org/abs/2307.07870}{{2307.07870}}}
\end{botherref}
\endbibitem

%%% 30
\bibitem[\protect\citeauthoryear{Ke et~al.}{2024}]{ke2024exploringfrontiersllmspsychological}
\begin{botherref}
\oauthor{\bsnm{Ke}, \binits{L.}},
\oauthor{\bsnm{Tong}, \binits{S.}},
\oauthor{\bsnm{Cheng}, \binits{P.}},
\oauthor{\bsnm{Peng}, \binits{K.}}:
Exploring the Frontiers of LLMs in Psychological Applications: A Comprehensive Review
(2024).
\url{https://arxiv.org/abs/2401.01519}
\end{botherref}
\endbibitem

%%% 31
\bibitem[\protect\citeauthoryear{Demszky et~al.}{2023}]{demszky2023using}
\begin{barticle}
\bauthor{\bsnm{Demszky}, \binits{D.}},
\bauthor{\bsnm{Yang}, \binits{D.}},
\bauthor{\bsnm{Yeager}, \binits{D.S.}},
\bauthor{\bsnm{Bryan}, \binits{C.J.}},
\bauthor{\bsnm{Clapper}, \binits{M.}},
\bauthor{\bsnm{Chandhok}, \binits{S.}},
\bauthor{\bsnm{Eichstaedt}, \binits{J.C.}},
\bauthor{\bsnm{Hecht}, \binits{C.}},
\bauthor{\bsnm{Jamieson}, \binits{J.}},
\bauthor{\bsnm{Johnson}, \binits{M.}}, \betal:
\batitle{Using large language models in psychology}.
\bjtitle{Nature Reviews Psychology}
\bvolume{2}(\bissue{11}),
\bfpage{688}--\blpage{701}
(\byear{2023})
\end{barticle}
\endbibitem

%%% 32
\bibitem[\protect\citeauthoryear{Almeida et~al.}{2024}]{ALMEIDA2024104145}
\begin{barticle}
\bauthor{\bsnm{Almeida}, \binits{G.F.C.F.}},
\bauthor{\bsnm{Nunes}, \binits{J.L.}},
\bauthor{\bsnm{Engelmann}, \binits{N.}},
\bauthor{\bsnm{Wiegmann}, \binits{A.}},
\bauthor{\bsnm{Araújo}, \binits{M.}}:
\batitle{Exploring the psychology of llms’ moral and legal reasoning}.
\bjtitle{Artificial Intelligence}
\bvolume{333},
\bfpage{104145}
(\byear{2024})
\doiurl{10.1016/j.artint.2024.104145}
\end{barticle}
\endbibitem

%%% 33
\bibitem[\protect\citeauthoryear{Zhang et~al.}{2023}]{zhang2023exploring}
\begin{botherref}
\oauthor{\bsnm{Zhang}, \binits{J.}},
\oauthor{\bsnm{Xu}, \binits{X.}},
\oauthor{\bsnm{Deng}, \binits{S.}}:
Exploring collaboration mechanisms for llm agents: A social psychology view.
arXiv preprint arXiv:2310.02124
(2023)
\end{botherref}
\endbibitem

%%% 34
\bibitem[\protect\citeauthoryear{Hu et~al.}{2024}]{hu2024psycollmenhancingllmpsychological}
\begin{botherref}
\oauthor{\bsnm{Hu}, \binits{J.}},
\oauthor{\bsnm{Dong}, \binits{T.}},
\oauthor{\bsnm{Gang}, \binits{L.}},
\oauthor{\bsnm{Ma}, \binits{H.}},
\oauthor{\bsnm{Zou}, \binits{P.}},
\oauthor{\bsnm{Sun}, \binits{X.}},
\oauthor{\bsnm{Guo}, \binits{D.}},
\oauthor{\bsnm{Wang}, \binits{M.}}:
PsycoLLM: Enhancing LLM for Psychological Understanding and Evaluation
(2024).
\url{https://arxiv.org/abs/2407.05721}
\end{botherref}
\endbibitem

%%% 35
\bibitem[\protect\citeauthoryear{Lai et~al.}{2023}]{lai2023psy}
\begin{botherref}
\oauthor{\bsnm{Lai}, \binits{T.}},
\oauthor{\bsnm{Shi}, \binits{Y.}},
\oauthor{\bsnm{Du}, \binits{Z.}},
\oauthor{\bsnm{Wu}, \binits{J.}},
\oauthor{\bsnm{Fu}, \binits{K.}},
\oauthor{\bsnm{Dou}, \binits{Y.}},
\oauthor{\bsnm{Wang}, \binits{Z.}}:
Psy-llm: Scaling up global mental health psychological services with ai-based large language models.
arXiv preprint arXiv:2307.11991
(2023)
\end{botherref}
\endbibitem

%%% 36
\bibitem[\protect\citeauthoryear{Bill and Eriksson}{2023}]{bill2023fine}
\begin{botherref}
\oauthor{\bsnm{Bill}, \binits{D.}},
\oauthor{\bsnm{Eriksson}, \binits{T.}}:
Fine-tuning a LLM using Reinforcement Learning from Human Feedback for a Therapy Chatbot Application.
Dissertation
(2023)
\end{botherref}
\endbibitem

%%% 37
\bibitem[\protect\citeauthoryear{Wang et~al.}{2024}]{wang2024towards}
\begin{botherref}
\oauthor{\bsnm{Wang}, \binits{J.}},
\oauthor{\bsnm{Xiao}, \binits{Y.}},
\oauthor{\bsnm{Li}, \binits{Y.}},
\oauthor{\bsnm{Song}, \binits{C.}},
\oauthor{\bsnm{Xu}, \binits{C.}},
\oauthor{\bsnm{Tan}, \binits{C.}},
\oauthor{\bsnm{Li}, \binits{W.}}:
Towards a client-centered assessment of llm therapists by client simulation.
arXiv preprint arXiv:2406.12266
(2024)
\end{botherref}
\endbibitem

%%% 38
\bibitem[\protect\citeauthoryear{Kang}{2016}]{kang2016fruit}
\begin{botherref}
\oauthor{\bsnm{Kang}, \binits{H.}}:
The fruit of my woman.
trans. D. Smith (London: Granta, 2000),
2--24
(2016)
\end{botherref}
\endbibitem

%%% 39
\bibitem[\protect\citeauthoryear{Fosse}{2018}]{fosse2018scenes}
\begin{bbook}
\bauthor{\bsnm{Fosse}, \binits{J.}}:
\bbtitle{Scenes from a Childhood}.
\bpublisher{Fitzcarraldo Editions},
\blocation{London}
(\byear{2018})
\end{bbook}
\endbibitem

%%% 40
\bibitem[\protect\citeauthoryear{Liu et~al.}{2023}]{liu2023transformation}
\begin{botherref}
\oauthor{\bsnm{Liu}, \binits{Z.}},
\oauthor{\bsnm{Li}, \binits{Y.}},
\oauthor{\bsnm{Cao}, \binits{Q.}},
\oauthor{\bsnm{Chen}, \binits{J.}},
\oauthor{\bsnm{Yang}, \binits{T.}},
\oauthor{\bsnm{Wu}, \binits{Z.}},
\oauthor{\bsnm{Hale}, \binits{J.}},
\oauthor{\bsnm{Gibbs}, \binits{J.}},
\oauthor{\bsnm{Rasheed}, \binits{K.}},
\oauthor{\bsnm{Liu}, \binits{N.}}, et al.:
Transformation vs tradition: Artificial general intelligence (agi) for arts and humanities.
arXiv preprint arXiv:2310.19626
(2023)
\end{botherref}
\endbibitem

\end{thebibliography}

\end{document}